# Solving MAP Exactly using Systematic Search

James D. Park and Adnan Darwiche
Computer Science Department
University of California
Los Angeles, CA 90095
{jd,darwiche}@cs.ucla.edu

## Abstract

MAP is the problem of finding a most probable instantiation of a set of variables in a Bayesian network, given some partial evidence about the complement of that set. Unlike posterior probabilities, or MPE (a special case of MAP), the time and space complexity of structure-based algorithms for MAP are *not only* exponential in the network treewidth, but in a larger parameter known as the *constrained* treewidth. In practice, this means that computing MAP can be orders of magnitude more expensive than computing posterior probabilities or MPE. We introduce in this paper a new, simple upper bound on the probability of a MAP solution, which is shown to be generally much tighter than existing bounds. We then use the proposed upper bound to develop a branch–and–bound search algorithm for solving MAP exactly. Experimental results demonstrate that the search algorithm is able to solve many problems that are far beyond the reach of any structure–based method for MAP. For example, we show that the proposed algorithm can compute MAP exactly and efficiently for some networks whose constrained treewidth is more than 40.

## 1 Introduction

The problem of finding the Maximum a Posteriori hypothesis (MAP) is to find the most likely configuration of a set of variables given some evidence.

One specialization of MAP which has received a lot of attention is the Most Probable Explanation (MPE). MPE is the problem of finding the most likely configuration of a set of variables given *complete* evidence about the complement of that set. The primary reason for this attention is that MPE is much easier than its MAP generalization. Specifically, the decision problem for MPE is NP–complete while the corresponding MAP problem is NP$^{PP}$–complete [7]. Unfortunately, MPE is not always suitable for the task of providing explanations.

Mainstream, structure-based methods can solve MPE in time which is exponential in treewidth [3, 10, 6, 1, 5]. These methods can also solve MAP, but they are exponential in the constrained treewidth, which can be much larger than treewidth, often pushing the problem beyond the threshold of feasibility [8, 7]. To deal with the computational intractability of MAP, recent approaches have focused on best–effort approximations which are only exponential in treewdith (as opposed to constrained treewidth) [2, 8, 7]. Although such algorithms show much promise, they cannot guarantee the optimality of solutions they provide.

The focus of this paper is then on providing a systematic search algorithm for solving MAP *exactly*. The presented algorithm is a depth–first branch–and–bound algorithm, which hinges on a new method for producing an upper bound on the probability of a MAP solution. We therefore study this new upper bound extensively and compare it to the main other bound available, which is based on mini–buckets [4]. We note that the complexity of our proposed bound is exponential in treewidth. Hence, our proposed algorithm is useful on networks for which solving MPE is considered tractable. This is similar to the approach presented in [8], except that our proposed method guarantees the optimality of the obtained solution, while the approach of [8] cannot offer such a guarantee.

This paper is structured as follows. We start in Section 2 by a review of variable elimination for solving MAP and MPE, which provides a context for introducing some technical preliminaries that we need in the rest of the paper. We then introduce in Section 3 a new upper bound on the probability of a MAP solution, and compare it theoretically and experimentally

---

**Algorithm 1** VE($\Phi, \mathbf{e}, \mathbf{S}, \mathbf{M}$): returns a probability.

1: $\pi$ := an ordering of the variables in $\mathbf{S}, \mathbf{M}$ where variables $\mathbf{M}$ are last in the order
2: $\Psi$ := the set of potentials $\phi_\mathbf{e}$, where $\phi \in \Phi$
3: **for** each variable $V$ in $\pi$ in order **do**
4:     remove from $\Psi$ the potentials mentioning variable $V$ and multiply them to form potential $\psi_V$
5:     **if** $V$ is a variable in $\mathbf{M}$ **then**
6:         add $\max_V \psi_V$ to $\Phi$
7:     **else**
8:         add $\sum_V \psi_V$ to $\Phi$
9: $\psi$ := the single potential in $\Psi$
10: return the single number assigned by $\psi$

---

to the bound based on mini–buckets. In Section 4, we present a systematic search method for solving MAP exactly, in which the upper bound algorithm plays a central role. Section 5 is then dedicated to experimental results demonstrating the effectiveness of the proposed algorithm. Section 6 concludes with a review of the primary results of the paper. Proofs of all theorems appear in the appendix.

## 2 Variable Elimination for MAP

We will review the algorithm of variable elimination in this section for computing MAP and MPE [3, 10].

Given a Bayesian network, we will treat each conditional probability table (CPT) $\phi$ as a potential over its corresponding variables $\mathbf{X}$; that is, a mapping from the instantiations $\mathbf{x}$ of these variables into the interval $[0, 1]$. Moreover, given a potential $\phi$, and an instantiation $\mathbf{e}$ of some network variables $\mathbf{E}$, we will use the notation $\phi_\mathbf{e}$ to indicate a new potential in which we fix the variables of $\phi$ to their corresponding values in $\mathbf{e}$, and then drop out the instantiated variables. That is, if $\mathbf{X} = \mathbf{YZ}$ are the variables of $\phi$, and $\mathbf{Z}$ are those appearing in instantiation $\mathbf{e}$, then $\phi_\mathbf{e}$ is a potential over variables $\mathbf{Y}$ and $\phi_\mathbf{e}(\mathbf{y}) = \phi(\mathbf{yz})$, where $\mathbf{z}$ are the values of variables $\mathbf{X}$ in instantiation $\mathbf{e}$.

We will also assume familiarity with the usual operations on potentials, including the multiplication of two potentials $\phi$ and $\psi$, denoted $\phi\psi$; the summing–out of variables $\mathbf{X}$ from potential $\phi$, denoted $\sum_\mathbf{X} \phi$; and the maxing–out of variables $\mathbf{X}$, denoted $\max_\mathbf{X} \phi$. A potential over the empty set of variables is called *trivial* and assigns a single number to the empty instantiation.

Given a Bayesian network with CPTs $\Phi$, let the network variables be partitioned into three sets: $\mathbf{E}, \mathbf{S}$ and $\mathbf{M}$. Here, $\mathbf{E}$ is the set of variables whose values are known, $\mathbf{S}$ is the set of variables that we want to sum out, and $\mathbf{M}$ is the set of variables that we want to maximize over. Given an instantiation $\mathbf{e}$ of variables $\mathbf{E}$, the *MPE problem* is that of finding an instantiation $\mathbf{s}, \mathbf{m}$ of variables $\mathbf{S}, \mathbf{M}$ which maximizes the probability of $\mathbf{s}, \mathbf{m}, \mathbf{e}$. The *MAP problem* is that of finding and instantiation $\mathbf{m}$ of variables $\mathbf{M}$ which maximizes the probability of $\mathbf{m}, \mathbf{e}$. Finally, the *PR problem* is that finding the probability of instantiation $\mathbf{e}$. Algorithm 1 provides pseudocode for a generic variable elimination algorithm which can be used to solve all of these problems. Specifically, the algorithm computes the following trivial potential:

$$\psi = \max_\mathbf{M} \sum_\mathbf{S} \prod_{\phi \in \Phi} \phi_\mathbf{e},$$

which assigns a single number $p$ to the empty instantiation. If $\mathbf{S}$ is empty, then $p$ is the probability of an MPE solution. If $\mathbf{M}$ is empty, then $p$ is the probability of evidence $\mathbf{e}$. Otherwise, $p$ is the probability of a MAP solution. The MPE/MAP solutions can be recovered via some bookkeeping that we omit for simplicity.

We will define the width of elimination order $\pi$ used on Line 1 of Algorithm 1 as $\log_2 s - 1$, where $s$ is the size of the largest potential $\psi_V$ produced on Line 4. Here, the size $s$ is defined as the number of instantiations in the domain of $\psi_V$. Variable elimination is linear in the number of network variables, and exponential only in the width of used elimination order. We note here that our definition of width is sensitive to the cardinalities of variables, and deviates from the classical definition which ignores such cardinalities. We adopt this definition since some of the networks we experiment with later have variables with cardinalities of over 60 and we need the definition to truly reflect the practical difficulties when reasoning with such networks. We also note that our definition corresponds to the classical definition when all variables are binary.

A key observation about the above variable elimination algorithm is that when $\mathbf{S}$ is empty (i.e., we have an MPE problem), then any variable order $\pi$ can be used. The same is true when $\mathbf{M}$ is empty (i.e., we have a PR problem). But when neither set of variables is empty (a MAP problem), the order $\pi$ is *constrained*. The width of best, unconstrained variable order is known as the *treewidth* of given network.[1] The width of best, constrained variable order is known as the *constrained treewidth*. In practice, the constrained treewidth can be much higher than treewidth, pushing some problems beyond the limit of feasibility [8]. For example, variable elimination for some MAP queries on polytrees require exponential time, while MPE and PR queries can be computed in linear time [7].

---

[1]Technically speaking, this requires the set of evidence variables $\mathbf{E}$ to be empty, otherwise, treewidth would be a function of both network and evidence variables.

The following theorem explains the need for a constrained order when solving MAP, and is key to a method we present in the following section for producing an upper bound on the probability of MAP.

**Theorem 1** *Let $\phi$ be a potential over disjoint variables $X, Y, \mathbf{Z}$. Then*

1. $\sum_X \sum_Y \phi = \sum_Y \sum_X \phi$.

2. $\max_X \max_Y \phi = \max_Y \max_X \phi$.

3. $[\sum_X \max_Y \phi](\mathbf{z}) \geq [\max_Y \sum_X \phi](\mathbf{z})$ *for all instantiations $\mathbf{z}$ of variables $\mathbf{Z}$. Moreover, the equality holds only when there is some value $y$ of variable $Y$ such that $\max_Y \phi_x = \phi_{xy}$ for all values $x$ of variable $X$. That is, the optimal value of variable $Y$ is independent of variable $X$.*

So, while summation commutes with summation, and maximization commutes with maximization, summation does not commute with maximization. This means that when solving PR, the summations can be arbitrarily permuted, allowing the elimination order to be chosen so as to minimize the width. The same is true for MPE, but this ceases to hold for MAP where the order used must be selected from among the orders in which all summation variables are eliminated before any maximization variables. There are sometimes orders that interleave summation and maximization variables which are still valid, but for any such order, there is a non–interleaved order of the same width [7].

As we show in the next section, even though the use of an arbitrary elimination order for MAP can produce an incorrect result, such use can be very useful for search–based algorithms as it is guaranteed to produce an *upper bound* on the correct result.

## 3 A New Upper Bound on the Probability of MAP

The goal of this section is to provide a new upper bound on the probability of a MAP solution, in addition to some important techniques for computing this bound effectively. We will then use the proposed bound in the next section to devise a branch–and–bound algorithm for computing MAP exactly. In describing our results, we will use MAP($\mathbf{M}, \mathbf{e}$) to denote a MAP problem with maximization variables $\mathbf{M}$, and evidence $\mathbf{e}$, and BD($\mathbf{M}, \mathbf{e}$) to denote an upper bound on the probability of a MAP solution for problem MAP($\mathbf{M}, \mathbf{e}$).

Our upper bound is based on the following result. We can use any elimination order on Line 1 of Algorithm 1, in which case the returned number will be an upper bound on the probability of a MAP solution. This result follows from Theorem 1 and the following observation. First, notice that from any elimination order $\pi$, a valid MAP order $\pi'$ can be produced by successively commuting a maximization variable $M \in \mathbf{M}$ with a summation variable $S \in \mathbf{S}$, whenever variable $M$ is immediately before the summation variable $S$ in the order. For example, consider maximization variables $X, Y$ and $Z$, summation variables $A, B$ and $C$, and the order $\pi = AXYBZC$. This is not a valid order for MAP. Yet, we can convert this order to the valid order $\pi' = ABCXYZ$ using the following steps: $AXYBCZ$ $\xrightarrow{YB}_{BY}$ $AXBYCZ$ $\xrightarrow{YC}_{CY}$ $AXBCYZ$ $\xrightarrow{XB}_{BX}$ $ABXCYZ$ $\xrightarrow{XC}_{CX}$ $ABCXYZ$.

Given Theorem 1, and Algorithm 1, each of the above orders will produce a number which is guaranteed to be no less than the number produced by the following order in the sequence. Hence, if we use the invalid order $\pi$ instead of the valid order $\pi'$, we are guaranteed to obtain a number which is an upper bound on the probability of a MAP solution. We actually have the following stronger result.

**Theorem 2** *If on Line 1 of Algorithm 1 we use an arbitrary order of variables $\mathbf{S}, \mathbf{M}$, the number $p$ returned by the algorithm obeys $\Pr(\mathbf{m}, \mathbf{e}) \leq p \leq \Pr(\mathbf{e})$, where $\mathbf{m}$ is a MAP solution.*

Recall that if we are allowed to use an arbitrary elimination order in Algorithm 1, the complexity of this variable elimination algorithm will be exponential in treewidth instead of constrained treewidth. Hence, Theorem 2 allows us to compute an upper bound on the probability of a MAP solution for problems whose constrained treewidth is too large, yet whose treewidth is acceptable. This includes networks for which MPE and PR can be solved practically using variable elimination.

As it turns out, even though any arbitrary variable order can be used to produce an upper bound, some orders will produce tighter upper bounds than others. Intuitively, the closer the used order is to a valid order, the tighter the bound is expected to be. We will next discuss a technique for selecting one of the better (invalid) orders. The technique will also be helpful in producing additional information which will be used by our branch–and–bound algorithm to be discussed in the following section.

### 3.1 Computing the Upper Bound using Jointree Algorithms

The upper bound we discussed earlier can be computed using the jointree algorithm, since the choice of a root in the jointree, and directing messages towards that root, corresponds to the application of variable elimination using a particular order which can be extracted from the jointree and chosen root. We will indeed use the jointree algorithm for computing the upper bound, instead of classical variable elimination, for a number of reasons that will become apparent later.

In particular, we will adapt the Shenoy–Shafer algorithm, which is one variant of the jointree algorithm [9]. We start with a jointree for the given network with potentials $\Phi$, and then assign each network potential $\phi \in \Phi$ to a jointree cluster $i$ containing the variables of that potential. Given evidence $\mathbf{e}$, summation variables $\mathbf{S}$, and maximization variables $\mathbf{M}$, let $\Phi_i$ be the product of all $\phi_{\mathbf{e}}$, where $\phi$ is a potential assigned to cluster $i$. We will define the message sent from cluster $i$ to neighboring cluster $j$ as follows:

$$M_{ij} = \max_{\mathbf{X}} \sum_{\mathbf{Y}} \Phi_i \prod_{k \neq j} M_{ki},$$

where $\mathbf{X} \subseteq \mathbf{M}$ and $\mathbf{Y} \subseteq \mathbf{S}$ are all variables that appear in cluster $i$ but not in cluster $j$.

The resulting jointree algorithm has the following semantics.

**Theorem 3** *For any cluster $i$ with maximization variables $\mathbf{X} \subseteq \mathbf{M}$ and summation variables $\mathbf{Y} \subseteq \mathbf{S}$, the potential*

$$\max_{\mathbf{X}} \sum_{\mathbf{Y}} \Phi_i \prod_{k} M_{ki}$$

*is trivial and contains an upper bound on the probability of a MAP solution, $BD(\mathbf{M}, \mathbf{e})$. Moreover, for any variable $X \in \mathbf{X}$, the upper bounds $BD(\mathbf{M} \setminus X, \mathbf{e}x)$ for all values $x$ of variable $X$ are available in the following potential:*

$$\max_{\mathbf{X}-\{X\}} \sum_{\mathbf{Y}} \Phi_i \prod_{k} M_{ki}.$$

Given the above theorem, one can compute an upper bound by choosing any cluster $i$ as the root, and directing messages towards that root. This corresponds to what is known as the *inward pass* of a jointree algorithm. If we follow this pass with the *outward pass*, in which messages are directed away from the root, we can also compute all upper bounds of the form $BD(\mathbf{M} \setminus X, \mathbf{e}x)$ simultaneously. These upper bounds will be especially useful in the search–based algorithm for MAP to be discussed in the following section.

Our search–based algorithm will also need to compute upper bounds of the form $BD(\mathbf{M} \setminus \mathbf{X}, \mathbf{e}x)$, where $\mathbf{X}$ is a set of variables instead of a singleton, as given above. For this, we have no choice but to assert further evidence $\mathbf{x}$ on the jointree and to re-apply the inward/outward passes described above. We will later use $assert(X = x)$ and $assert(X \neq x)$ to denote the classical process of setting evidence on a jointree.

## 3.2 Improving the Bound Quality

Given a jointree $T$ with a largest cluster of size $s$, we will now discuss a technique for generating another jointree $T'$ whose largest cluster has a size no greater than $s$, yet will produce tighter upper bounds.

The idea is to select some root cluster $r$ in jointree $T$, and migrate maximization variables $\mathbf{M}$ toward the root, therefore, inducing an elimination order that is closer to a valid order. Specifically, let $k$ and $i$ be two clusters in the jointree, where $i$ is closer to the root $r$ (we call $k$ the *child* of $i$ in this case). Any maximization variable $V$ that appears in cluster $k$ is added to cluster $i$ as long as the new size of cluster $i$ does not exceed $s$. Algorithm 2 gives pseudocode for a recursive algorithm to promote maximization variables $\mathbf{M}$ towards the root $r$ in jointree $T$, without increasing the size of its clusters beyond $s$. The algorithm should initially be called with promote$(r, T, r, \mathbf{M}, s)$.

---

**Algorithm 2** promote$(i, T, r, \mathbf{M}, s)$: Modifies jointree $T$ while keeping the size of its clusters $\leq s$.

    **for** each child cluster $k$ of cluster $i$ **do**
        call promote$(k, T, r, \mathbf{M}, s)$
        $\mathbf{V} :=$ all variables $V \in \mathbf{M}$ that appear in a child cluster $k$ of $i$ but not in $i$
        **while** $\mathbf{V} \neq \emptyset$ and the size of cluster $i$ is $\leq s$ **do**
            remove a member $V$ from $\mathbf{V}$
            **if** adding $V$ to cluster $i$ keeps its size $\leq s$, add variable $V$ to cluster $i$

---

Figure 1 illustrates a jointree before and after promotion of the maximization variables.

Effectively, each time a variable is promoted from cluster $k$ to cluster $i$ (which is closer to the root $r$), the maximization over that variable is postponed, pushing it past all of the summation variables that are eliminated in cluster $i$ or its children. This has a monotonic effect on the upper bound, although in rare cases, the bound may remain the same. We stress here that the promotion technique is meant to improve the bound computed by directing messages towards the root $r$. After such promotion, the quality of a bound computed by directing messages to some other root $r'$ may actually worsen. More on this later.

We now describe another technique for improving the quality of the computed upper bound. Specifically, sometimes a Bayesian network has variables with a large number of states. For example, the Munin2 network discussed later has a variable with 21 states. When the domain of a variable is large, promoting it increases significantly the size of some cluster. When many of the maximization variables are large, this often allows for relatively few possible promotions. To relieve this to some extent, we replace each variable with variables whose sizes are the prime factors of the original size. For example, a potential $\phi$ over a variable $X$ which has 21 possible values is replaced with a potential $\phi'$, over variables $X_1$ and $X_2$ which have sizes 3 and 7 respectively. There is a one-to-one correspondence between instances of the new variables, and the old, so converting the potentials is straightforward. This conversion in many cases allows promotion of some of the factored variables in cases where the original variable would not be promoted.[2]

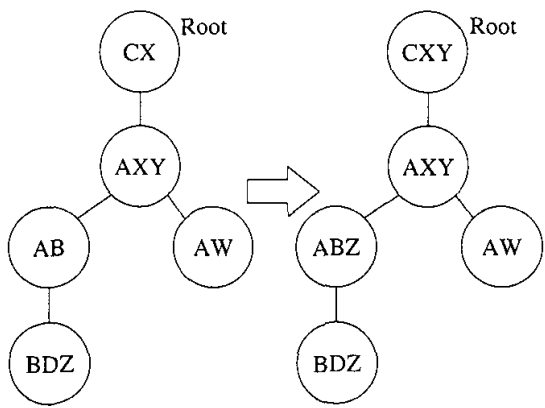

Figure 1: The original jointree, and the jointree produced by promoting the maximization variables $W, X, Y,$ and $Z$ while preserving the size of the largest cluster $s$ at 8 (all variables are binary).

### 3.3 The Mini–Bucket Upper Bound

The only other approach we are aware of for providing an upper bound on MAP is the one based on mini-buckets [4]. To explain this bound, note that the complexity of variable elimination is dominated by the size of potential $\psi_V$ created on Line 4 of Algorithm 1. The idea behind the mini-bucket method to control the size of this potential using a parameter $k$ which puts a limit on the size of $\psi_V$. Specifically, if the potential size is to exceed $k$, the mini-bucket method will not create $\psi_V$, but an approximate version of it, $\psi'_V$. We will not discuss the specific method for this approximation, but suffice it to say that the quality of the approximation improves as $k$ increases. Moreover, if $k$ is large enough, the method converges to the classical variable elimination algorithm, giving us the correct result. The mini–bucket method thus allows a smooth transition between accuracy and efficiency based on the supplied complexity parameter $k$. Note that our bound admits such a tradeoff, to some extent, as we can control the width through the choice of which invalid order we use, obtaining a potentially exact result when the width equals the constrained treewidth. However, our method is not as flexible as mini-buckets in that the complexity of our method is exponential at least in the treewidth, while mini-buckets allows the the complexity parameter $k$ to be chosen arbitrarily.

To compare the quality of the two upper bounds we ran experiments with randomly–generated and real-world networks. Specifically, we generated 100 MAP problems from random Bayesian networks and applied both upper bounds to each problem. The networks were generated using the method detailed in [1]. Each network consists of 100 binary variables and was generated using connectively parameter 20 (this tends to produce networks with widths of about 20). We set evidence on the leaves, and randomly selected 25 MAP variables. We approximated the treewidth of the network using the min–fill heuristic, and used that as the complexity parameter for both approximation methods, and recorded the upper bounds they produced. Since both methods produce upper bounds, the better approximation is the smaller one. Our method ranged from $9.3 \times 10^4$ to $1.5 \times 10^9$ times smaller than the upper bound produced using mini-buckets. We performed a similar experiment on the real–world Bayesian networks which appear in the experimental results section (Table 3). For each network, we performed 10 experiments. In each experiment we randomly selected one quarter of the variables as MAP variables, and set evidence on the leaves, taking care to ensure probability of the evidence remained positive. Table 1 provides statistics on the relative performance for each network. Our method produced bounds that were at least $2.4 \times 10^5$ times smaller than the mini-bucket bounds in all of the experiments, except for those on the Water network, where it was at least $2.6 \times 10^3$ times smaller. In fact, in contrast to our bound, which is always bounded above by the probability of evidence, some of the bounds produced by the mini–bucket method were greater than one, an thus completely uninformative.

## 4　Solving MAP using Systematic Search

We present in this section an algorithm for solving MAP exactly using systematic search. Specifically, given a Bayesian network, with evidence **e** and MAP variables $\mathbf{M}$, our goal is to identify an instantiation $\mathbf{m}$ of variables $\mathbf{M}$ which maximizes the probability of $\mathbf{m}, \mathbf{e}$.

---

[2]This is not always a satisfactory solution. For example, the Barley network has a variable with 67 values. As 67 is prime, this method does not help. Methods which grow the value to a more easily factored value (for example from 67 to 72 say) suggest themselves, but we have not explored them. Such a technique will typically increase the treewidth as well.

| Network | Min | Max |
|---|---|---|
| Barley | $2.4 \times 10^5$ | $3.2 \times 10^{12}$ |
| Diabetes | $1.0 \times 85$ | $4.0 \times 10^{111}$ |
| Mildew | $1.7 \times 10^6$ | $2.7 \times 10^8$ |
| Munin2 | $1.6 \times 10^{122}$ | $1.3 \times 10^{142}$ |
| Munin3 | $1.9 \times 10^{124}$ | $4.0 \times 10^{152}$ |
| Munin4 | $7.9 \times 10^{133}$ | $3.8 \times 10^{151}$ |
| Pigs | $3.6 \times 10^{36}$ | $1.1 \times 10^{45}$ |
| Water | $2.6 \times 10^3$ | $2.1 \times 10^9$ |

Table 1: Statistics on the ratio of the mini-bucket bound to our bound. Notice that in all experiments our bound was significantly tighter.

As MAP is a discrete optimization problem, it can be solved by systematically searching for the optimal instantiation $\mathbf{m}$ of the MAP variables $\mathbf{M}$. Moreover, the upper bound on the probability of a MAP solution can be used as the basis for a depth–first branch–and–bound algorithm for computing an exact solution.

Specifically, the nodes in the search tree represent partial instantiations of the MAP variables $\mathbf{M}$. The root node corresponds to the empty instantiation. The children of a node which represents instantiation $\mathbf{x}$, where $\mathbf{X} \subseteq \mathbf{M}$, are nodes which represent instantiations $\mathbf{x}, v$ for some variable $V$ in $\mathbf{M} \setminus \mathbf{X}$. Leaves of the search tree correspond to different instantiations $\mathbf{m}$ of MAP variables $\mathbf{M}$. Since each node in the search tree corresponds to a variable instantiation, we will identify that node by the corresponding instantiation.

We associate a *score* with each leaf node $\mathbf{m}$ in the search tree, which is equal to $\Pr(\mathbf{m}, \mathbf{e})$. Hence, for an internal node $\mathbf{x}$, the best leaf node below $\mathbf{x}$ (one with highest score) is a solution to $\text{MAP}(\mathbf{M} \setminus \mathbf{X}, \mathbf{ex})$. The basic idea of the search algorithm is to perform a depth–first search on the tree, while computing an upper bound $\text{BD}(\mathbf{M} \setminus \mathbf{X}, \mathbf{ex})$ at each internal node $\mathbf{x}$. If the upper bound is less than or equal to the score of the best leaf encountered so far, the children of node $\mathbf{x}$ are not explored since none of the leaves below $\mathbf{x}$ can improve the existing solution.

While the general idea of the algorithm is very simple, there are a number of optimizations that are needed in order to significantly improve the search performance. Several of the optimizations are based on the second part of Theorem 3 and are discussed next.

**Variable Ordering.** One important technique is dynamic variable ordering. At each internal node $\mathbf{x}$, we need to choose a variable $V \in \mathbf{M} \setminus \mathbf{X}$ to instantiate next, hence, generating the children of node $\mathbf{x}$. The specific variable chosen can have a dramatic effect on the efficiency of the search. We experimented with a variety of variable selection heuristics. The one we found most effective works as follows. The first step is to compute for each potential variable $V \in \mathbf{M} \setminus \mathbf{X}$ the bound $B_v = \text{BD}(\mathbf{M} \setminus (\mathbf{X} \cup \{V\}), \mathbf{ex}v)$ for each value $v$. This can be performed efficiently using the jointree algorithm presented in Section 3, by performing a local computation on a cluster than contains variable $V$ (see the second part of Theorem 3). Based on these bounds $B_v$, we choose a variable as follows:

1. Let $M_V = \max_v B_v$, that is, the best upper bound obtained for variable $V$.

2. Let $T_V = \sum_{B_v \geq b} B_v$, where $b$ is the score of the best leaf node visited so far. That is, $T_V$ is the sum of bounds for variable $V$ which are better than the best score $b$ obtained so far.

We then choose the variable $V$ that maximizes the ratio $M_V/T_V$. The reason we normalize, instead of choosing the variable $V$ with the largest $M_V$, is that for variables $V$ which appear in clusters farthest from the root, the upper bounds produced by the jointree algorithm tend to be artificially inflated. Recall that the promote operation tends to move most of the maximization variables toward the root. Although this improves the upper bound computed at the root, it degrades the bounds computed at other clusters; see Section 3.2.

**Value Ordering.** Another important technique is value ordering. When a variable is selected, its values are explored in decreasing order of their upper bounds. This has the effect of trying to search the instantiations deemed most likely to produce a better solution first, possibly increasing pruning later.

**Value Elimination.** Suppose that we are currently at node $\mathbf{x}$ in the search tree. Suppose further that for some value $v$ of a remaining variable $V$, we have that $B_v \leq b$, where $b$ is the score of best instantiation $\mathbf{m}$ we have so far. We clearly know at this stage that no instantiation compatible with $\mathbf{x}v$ should be explored, because regardless of how we instantiate the remaining variables in $\mathbf{M} \setminus (\mathbf{X} \cup \{V\})$, there is no way we can produce an instantiation which is better than the one we already have. Still, the mere presence of $v$ in the potentials can influence the bound. Specifically, even though the value $v$ cannot lead to an instantiation with a better score than $b$, the following may be possible. For some fixed instantiation of other variables in a potential $\phi$ containing variable $V$, the entry corresponding to $V = v$ in $\phi$ may have the largest value. Thus, when variable $V$ is maximized out, the value of some

**Algorithm 3** ComputeMAP($\Phi, \mathbf{M}, \mathbf{e}$): Returns an instantiation $\mathbf{m}$, which maximizes the probability of $\mathbf{e}, \mathbf{m}$, and its corresponding probability. $\Phi$ are the CPTs/potentials of a Bayesian network.

$\Psi :=$ all $\phi_\mathbf{e}$ where $\phi \in \Phi$
$T :=$ a jointree for potentials $\Phi$
$r :=$ a root in jointree $T$
$s :=$ size of largest cluster in $T$
promote($r, T, r, \mathbf{M}, s$)
$bSol :=$ an instantiation $\mathbf{m}$ of variables $\mathbf{M}$ obtained using sequential initialization and pure hill climbing
$bScore :=$ probability of instantiation $\mathbf{m}, \mathbf{e}$
Search($\mathbf{M},\{\}$)
Return $bSol$ and $bScore$

**Algorithm 4** Search($\mathbf{X},\mathbf{z}$): Sets the values of global variables $bSol$ and $bScore$.
1: $B := BD(\mathbf{X}, \mathbf{z})$ (jointree inward pass)
2: If $B \leq bScore$, then return
3: If $\mathbf{X} = \emptyset$, then $bScore := B$, $bSol := \mathbf{z}$, return
4: **for** each variable $V \in \mathbf{X}$ **do**
5:    **for** each value $v$ of $V$ **do**
6:       $B_v := BD(\mathbf{X} \setminus V, \mathbf{z}v)$ (jointree outward pass)
7: assert($V \neq v$) for each $v$ where $B_v \leq bScore$
8: $V :=$ variable in $\mathbf{X}$ that maximizes $M_V/T_V$
9: **for** values $v$ of $V$ in decreasing order of $B_v$ **do**
10:    If $B_v > bScore$, then assert($V = v$) and call Search($\mathbf{X} \setminus V, \mathbf{z}v$)
11: Retract the assertions performed on lines 6 and 10.

| Set | #/50 | | find | finish | c-w |
|---|---|---|---|---|---|
| Rand-30 | 49 | min | .2 | 0.6 | 27.0 |
| | | median | 1.3 | 11.9 | 33.0 |
| | | mean | 2.8 | 31.5 | 32.3 |
| | | max | 22.6 | 242.5 | 37.0 |
| Rand-40 | 43 | min | 0.1 | 2.1 | 25.0 |
| | | median | 3.6 | 26.7 | 40.0 |
| | | mean | 9.2 | 43.4 | 39.6 |
| | | max | 85.2 | 220.4 | 44.0 |

Table 2: The results for the random network problems. The columns report the number out of 50 that were solved, and statistics on the time to find the solution, the time to complete the search (both in *seconds*), and the constrained treewidth for the problems that completed.

of the entries in the resulting potential may include entries corresponding to $V = v$, needlessly loosening the bound. To prevent this, we assert the evidence $V \neq v$ on the underlying jointree, factoring this pruning information into the computation of our upper bound. In practice, this can have a significant effect on the bound.

**Initialization.** To initialize the search, we used a MAP approximation technique based on *local search*. In particular, we used the technique of hill climbing coupled with sequential initialization, as described in [8], but we omit the details for space limitations.

Algorithms 3 and 4 provide pseudocode of our systematic search algorithm for solving MAP exactly, which combines all of the techniques we discussed thus far.

## 5 Experimental Results

We applied the search algorithm to both randomly generated and real-world networks, and collected statistics about the computations.

We generated random networks using the method described in [1] while including 100 nodes in each network, and using a connectivity parameter of 20 (which tends to produce networks with treewidths of about 20). For the first set, we generated 50 networks, randomly selecting 30 MAP variables. We set as evidence the leaf nodes, taking care to ensure that the probability of the evidence was positive. The computation for each MAP problem was given 10 minutes to complete. For each problem instance we computed the time to encounter the solution, the time for the search to complete, and the approximate constrained treewidth, as computed by the min–fill heuristic.

One of the 50 problems did not complete in 10 minutes of computation. Statistics for the remaining 49 appear in Table 2 in the Rand-30 data set. The results reported are the number completed within the time limit, and the minimum, median, mean and maximum values for each of the following: the time to find the solution, the time to complete the search, and the approximate constrained treewidth. Next, we generated another set of results using the same experiments as before, except 40 variables were chosen to be maximization variables. 7 of the problems were not solved after 10 minutes of computation. Statistics on the remaining 43 problems appear in Table 2 labeled as the Rand-40 data set.

We also tested the search algorithm on real–world networks Barley, Diabetes, Mildew, Munin2, Munin3, Munin4, Pigs and Water. For each network, we ran 10 experiments. In each experiment, we selected the MAP variables randomly. Because the number of values per variable differs, we selected the MAP variables so that the state space of the MAP variables was about $2^{30}$. Specifically, we randomly selected network variables without replacement, adding the selected variable to the set of MAP variables if the total state space remained less than or equal to $2^{30}$, until no variables remained. We again instantiated the leaf nodes, ensuring that the evidence had positive probability. Each problem was given 10 minutes to run. Table 3 reports the number that completed, and contains statistics on the results of the runs that did complete.

The first thing to notice is that nearly all of the problems have an approximate constrained treewidths far too large to be computed using structure–based methods. In spite of the large widths, the search based algorithm was able to solve many of the problems relatively quickly.

Another thing to notice is the significant variability of solution time both for the random and the real–world networks. There are a number of factors that influence the difficulty, including the treewidth of the network (which influences the time per search step), the size of the prime factors of the variables (which limits the amount of possible promotion), the relative magnitude of the solution as compared to the non–solutions (which significantly affects pruning), and of couse luck in choosing the variable and value orderings so as to encounter the true solution early in the search. So, for example, problems for the Pigs network which have small treewidth, small variables and significant determinism proved much easier than those for Barley, which has large factor variables, and large treewidth.

Also, notice that the solution was often found far before algorithm was able to prove that that was the solution. Apart from this, the algorithm produces a candidate solution very quickly, and so can be used as an any time algorithm as well.

Overall, the search algorithm was able to solve the vast majority of the MAP problems we generated. The algorithm succeeds where structure–based methods cannot be applied because of the significant pruning afforded by the tightness of the upper bound.

| Network | #/10 |        | find  | finish | c-w  |
|---------|------|--------|-------|--------|------|
| Barley  | 3    | min    | 29.4  | 78.3   | 31.5 |
|         |      | median | 33.1  | 86.5   | 32.1 |
|         |      | mean   | 57.3  | 140.6  | 34.7 |
|         |      | max    | 109.5 | 257.0  | 40.4 |
| Diabetes| 6    | min    | 3.5   | 14.0   | 33.8 |
|         |      | median | 4.4   | 87.5   | 37.7 |
|         |      | mean   | 40.9  | 213.6  | 37.0 |
|         |      | max    | 222.6 | 592.8  | 40.1 |
| Mildew  | 10   | min    | 18.3  | 27.0   | 29.3 |
|         |      | median | 43.0  | 74.8   | 31.7 |
|         |      | mean   | 43.8  | 73.7   | 31.4 |
|         |      | max    | 85.8  | 114.1  | 34.2 |
| Munin2  | 10   | min    | 4.1   | 4.9    | 33.1 |
|         |      | median | 4.4   | 5.6    | 35.6 |
|         |      | mean   | 4.4   | 5.6    | 35.2 |
|         |      | max    | 5.0   | 6.1    | 36.1 |
| Munin3  | 10   | min    | 5.0   | 6.0    | 29.8 |
|         |      | median | 6.1   | 7.3    | 33.0 |
|         |      | mean   | 6.3   | 7.6    | 34.4 |
|         |      | max    | 8.0   | 9.8    | 42.0 |
| Munin4  | 10   | min    | 20.8  | 26.4   | 34.2 |
|         |      | median | 28.9  | 33.4   | 36.6 |
|         |      | mean   | 27.7  | 33.8   | 36.1 |
|         |      | max    | 35.6  | 48.0   | 37.1 |
| Pigs    | 10   | min    | 2.6   | 3.4    | 29.1 |
|         |      | median | 3.2   | 4.2    | 30.7 |
|         |      | mean   | 3.4   | 4.6    | 31.2 |
|         |      | max    | 4.6   | 7.0    | 37.0 |
| Water   | 10   | min    | 74.4  | 115.8  | 26.9 |
|         |      | median | 93.1  | 146.0  | 29.1 |
|         |      | mean   | 116.0 | 155.3  | 29.4 |
|         |      | max    | 215.3 | 223.2  | 32.5 |

Table 3: Results on real–world networks. The columns report the number out of 10 that were solved, and statistics on the time to find the solution, the time to complete the search (both in *seconds*), and the approximate constrained treewidth for the problems that completed.

## 6 Conclusion

We introduced a simple upper bound for MAP based on relaxing the elimination ordering constraint of variable elimination. Experimental results show that this bound is significantly tighter than previous MAP upper bounds. We used this upper bound as the basis for a systematic search algorithm for MAP. The search algorithm is able to solve many problems that are far beyond the reach of previous methods for solving MAP, significantly extending the class of MAP problems that can be solved efficiently.

## Acknowledgements

This work has been partially supported by NSF grant IIS-9988543 and MURI grant N00014-00-1-0617.

## Proof of Theorem 1

For each case, we will consider a function $\psi$, where $\psi_{xy} = \phi_{xyz}$ for a particular fixed $\mathbf{z}$. For part 1, because $X$ and $Y$ have a finite number of values, the expression on the left can be tranformed to the expression on the right, simply by rearranging the terms in the summation.

For part 2, as $X$ and $Y$ are finite, there are a finite number of values of the function $\psi$, and so there is a maximal value $m$. Then for some value $y$ of $Y$, $\max_X \psi_y = m$, and for any other value $y'$ of $Y$, $\max_X \psi_{y'} \leq m$. Thus, $\max_Y \max_X \psi = m$. By an analogous argument $\max_X \max_Y \psi = m$, so $\max_X \max_Y \psi = \max_Y \max_X \psi$.

For part 3, for any fixed $x$, $\max_Y \psi_x \geq \psi_{xy}$ for any $y$. So, summing over $X$ yields $\sum_X max_Y \psi \geq \sum_X \psi_y$. This is true for all values $y$ of $Y$, and so is true for the $y$ that maximizes $\sum_X \psi_y$. Hence $\sum_X \max_Y \psi \geq \max_Y \sum_X \psi$.

## Proof of Theorem 2

First, we prove a simple lemma that will aid in the proof of the theorem.

**Lemma 4** *Let $\Phi$ be a set of potentials, $\Phi_V$ be the potentials of $\Phi$ whose domain include $V$, and let $\Phi_o = \Phi \setminus \Phi_V$. Then $\sum_V \prod_{\phi \in \Phi} \phi = \phi_s \prod_{\phi \in \Phi_o} \phi$ where $\phi_s = \sum_V \prod_{\phi \in \Phi_V} \phi$. Similarly, $\max_V \prod_{\phi \in \Phi} = \phi_m \prod_{\phi \in \Phi_o} \phi$, where $\phi_m = \max_V \prod_{\phi \in \Phi_V}$.*

The proof is a simple process of factoring. Note that $\sum_V \prod_{\phi \in \Phi} \phi = \sum_V \left( \prod_{\phi \in \Phi_o} \phi \right) \left( \prod_{\phi \in \Phi_V} \phi \right)$. Now, since the first product does not mention $V$, it can be factored out of the summation, yielding $(\prod_{\phi \in \Phi_o} \phi) \sum_V \prod_{\phi \in \Phi_V} \phi = \phi_s \prod_{\phi \in \Phi_o} \phi$. Similar arguments apply to the maximization equation. ∎

Notice that Lemma 4 is just a statement about what happens during each step of variable elimination. Thus repeated application of Lemma 4 shows that performing variable elimination with order $\pi$ yields

$$O_{\pi(n)} O_{\pi(n-1)} \cdots O_{\pi(1)} \prod_{\phi \in \Phi} \phi$$

where $O_X = \sum_X$ if $X \in \mathbf{S}$, and $\max_X$ otherwise. Now, as discussed in the main text we can produce a valid order from any order by successively commuting adjacent summation and maximization variables, where the maximization variable appears first. By Theorem 1, each commutation yields an order that produces a value less than or equal to the previous. Thus, the value produced by any order is an upper bound on the MAP value.

For a non-negative function $f$, $\sum_X f \geq \max_X f$. Now, consider the expression generated by any order. Because the product of potentials is a non-negative function, we can replace each maximization with a summation over the same variable, and produce an expression that is greater or equal to the original expression. But this resulting expression yields $\Pr(\mathbf{e})$. Thus, the value produced by any order is bounded below by the true MAP probability, and bounded above by the probability of evidence. ∎

**Proof of Theorem 3**

First, we provide some simple semantics for the messages, then use that to prove the theorem.

**Lemma 5** *The message $M_{ij}$ from cluster $i$ to neighboring cluster $j$ is equivalent to $O_{\pi(n)} O_{\pi(n-1)} \ldots O_{\pi(1)} \prod_{\phi \in \Phi_{ij}} \phi$, for some order $\pi$ of the variables that appear in the $i$ side of the jointree, but not on the $j$ side, where $O_X = \sum_X$ if $X$ is a summation node, $\max_X$ otherwise, and $\Phi_{ij}$ is the set of potentials assigned to the $i$ side of the tree.*

The proof is by induction on the depth of the tree rooted at $i$, when the $j$ side of the tree is removed. For the base case when $i$ is a leaf, the message definition is $\max_{\mathbf{X}} \sum_{\mathbf{Y}} \phi_i$, where $\mathbf{X}$ and $\mathbf{Y}$ are the maximization and summation variables that appear in $i$, but not in $j$. So the lemma criteria is satisfied using any order where the variables $\mathbf{Y}$ are eliminated, then the the variables $\mathbf{X}$. Now, assume by way of induction that the lemma is satisfied for all messages whose subtrees are up to depth $k$. Now, consider a message from cluster $i$ to $j$ whose subtree has a depth of $k+1$. The message is defined by $\max_{\mathbf{X}} \max_{\mathbf{Y}} \phi_i \prod_{l \neq j} M_{li}$. By the inductive hypothesis, each message can be replaced by its corresponding expression, yielding $\max_{\mathbf{X}} \max_{\mathbf{Y}} \phi_i \prod_{l \neq j} O_{\pi_l(n_l)} O_{\pi_l(n_l-1)} \ldots O_{\pi_l(1)} \prod_{\phi \in \Phi_{li}} \phi$. Since any summation or maximization occurs only over variables that appear only in that message, the potentials can be multiplied and the operators for all of the messages can simply be concatenated. Similarly, as the variables of $\phi$ do not appear in any of the operators of the other messages, $\phi_i$ can be moved inside the innermost product. Choosing an order $\pi_{\mathbf{X}}$ and $\pi_{\mathbf{Y}}$, and letting $\pi$ be the concatenation $\pi_{l_1} \ldots \pi_{l_n} \pi_{\mathbf{Y}} \pi_{\mathbf{X}}$ of the orders corresponding to the child messages, and the variables eliminated from $i$ to $j$, we have $M_{ij} = O_{\pi(n)} O_{pi(n-1)} \ldots O_{\pi(1)} \prod_{\phi \in \Phi_{ij}} \phi$. This completes the inductive step. ∎

Now, for a cluster $i$, with maximization variables $\mathbf{X}$, and summation variables $\mathbf{Y}$, using the above Lemma, and using the same arguments it employed regarding pushing the multiplications inside, and concatenating the operators, we have $\max_{\mathbf{X}} \sum_{\mathbf{Y}} \phi_i \prod_k M_{ki} =$ $O_{\pi(n)} O_{\pi(n-1)} \ldots O_{\pi(1)} \prod_{\phi \in \Phi} \phi$ for some order $\pi$. As this is the same result as variable eliminatio with order $\pi$, it yields an upper bound on the MAP probability.

For the second part of the theorem, choose $\pi$ so that $X$ is the last variable eliminated. Before performing the last operation (maximizing out $X$), we have a potential over variable $X$, where each entry bounds the probability of the maximal instantiation of the other MAP variables, which is compatible with that value of $X$. This then is equivalent to a bound on the MAP probability for $MAP(\mathbf{M} \setminus \{X\}, xe)$. ∎